\documentclass[prodtf]{tlp_nodates}

\volume{\textbf{10} (4-6)}
\pubyear{2010}
\jdate{July}
\doi{S1471068410000153}
\pagerange{361--364}
\usepackage[hypertex]{hyperref}

\begin{document}

\title[Introduction to the 26th ICLP Special Issue]{Introduction to
  the 26th International Conference on Logic Programming Special
  Issue}

\author[Manuel Hermenegildo and Torsten Schaub]{MANUEL HERMENEGILDO\\
IMDEA Software Institute and U.\ Polit\'{e}cnica de Madrid (UPM), Spain\\
\email{manuel.hermenegildo@\{imdea.org|upm.es\}}\\
\and TORSTEN SCHAUB\\
University of Potsdam, Germany\\
\email{torsten@cs.uni-potsdam.de}}

\maketitle

% \begin{abstract}
% This is the abstract 
% \end{abstract}
\hrule

\ \\ [-2mm]

% \begin{keywords}
% \end{keywords}

The Logic Programming (LP) community, through the Association for
Logic Programming (ALP) and its Executive Committee, decided to
introduce for 2010 important changes in the way the main yearly
results in LP and related areas are published. Whereas such results
have appeared to date in standalone volumes of proceedings of the
yearly International Conferences on Logic Programming (ICLP), and this
method --fully in the tradition of Computer Science (CS)-- has served
the community well, it was felt that an effort needed to be made to
achieve a higher level of compatibility with the publishing mechanisms
of other fields outside CS.

% Since the first conference held in Marseilles in 1982, ICLP has been
% the premier international conference for presenting research in
% logic programming. ... List of previous conference locations??? Only
% in LIPIcs?

In order to achieve this goal without giving up the traditional CS
conference format a different model has been adopted starting in 2010
in which the yearly ICLP call for submissions takes the form of a
joint call for a) \emph{full papers} to be considered for publication
in a special issue of the journal, and b) shorter \emph{technical
  communications} to be considered for publication in a separate,
standalone volume, with both kinds of papers being presented by their
authors at the conference.  Together, the journal special issue and
the volume of short technical communications constitute the
\emph{proceedings} of ICLP.

This \emph{26th International Conference on Logic Programming Special
  Issue} is the first of a series of yearly special issues of Theory
and Practice of Logic Programming (TPLP) putting this new model into
practice. It contains the papers accepted from those submitted as full
papers (i.e., for TPLP) in the joint ICLP call for 2010.  The
collection of technical communications for 2010 appears in turn as
Volume 7 of the
\href{http://www.dagstuhl.de/en/publications/lipics}{Leibniz
  International Proceedings in Informatics (LIPIcs)} series, published
on line through the
\href{http://drops.dagstuhl.de/opus/institut_lipics.php?fakultaet=04}
{Dagstuhl Research Online Publication Server (DROPS).}
% A listing of these papers appears at the end of this issue.  
Both sets of papers were presented by their authors at this 26th ICLP.

Papers describing original, previously unpublished research and not
simultaneously submitted for publication elsewhere were solicited in
all areas of logic programming including but not restricted to:
\emph{Theory} (Semantic Foundations, Formalisms, Non-monotonic
Reasoning, Knowledge Representation), \emph{Implementation}
(Compilation, Memory Management, Virtual Machines, Parallelism),
\emph{Environments} (Program Analysis, Transformation, Validation,
Verification, Debugging, Profiling, Testing), \emph{Language Issues}
(Concurrency, Objects, Coordination, Mobility, Higher Order, Types,
Modes, Assertions, Programming Techniques), \emph{Related Paradigms}
(Abductive Logic Programming, Inductive Logic Programming, Constraint
Logic Programming, Answer-Set Programming), and \emph{Applications}
(Databases, Data Integration and Federation, Software Engineering,
Natural Language Processing, Web and Semantic Web, Agents, Artificial
Intelligence, Bioinformatics).
%
% \begin{description}
%   \itemsep=0pt
% \item[Theory:] Semantic Foundations, Formalisms, Non-monotonic Reasoning,
%   Knowledge Representation.
% \item[Implementation:] Compilation, Memory Management, Virtual Machines,
%   Parallelism.
% \item[Environments:] Program Analysis,  Transformation, Validation,
%   Verification, Debugging, Profiling, Testing.
% \item[Language Issues:] Concurrency, Objects, Coordination, Mobility,
%   Higher Order, Types, Modes, Assertions, Programming Techniques.
% \item[Related Paradigms:] Abductive Logic Programming, Inductive Logic
%   Programming, Constraint Logic Programming, Answer-Set Programming.
% \item[Applications:] Databases, Data Integration and Federation,
%   Software Engineering, Natural Language Processing, Web and Semantic
%   Web, Agents, Artificial Intelligence, Bioinformatics.
% \end{description}

% The full papers included also 
Special categories were \emph{application papers} (where the emphasis
was on their impact on the application domain) and \emph{system and
  tool papers} (where the emphasis was on the novelty, practicality,
usability and general availability of the systems and tools
described).  In the shorter \emph{technical communications} the
emphasis was on describing recent developments, new projects, and
other materials not yet ready for publication as full papers.  The
length limit for full papers was set at 15 pages plus bibliography for
full papers (approximately in line with the length of TPLP technical
notes) and for technical communications at 10 pages total.

In response to the call for papers 104 abstracts were received, 81 of
which remained finally as complete submissions. Of those, 69 were full
papers submitted to the TPLP special issue track (21 of them
applications or systems papers).  The program chairs acting as guest
editors organized the refereeing process with the help of the program
committee and numerous external reviewers.\footnote{The LIPIcs volume
  contains a complete list of referees.} Each paper was reviewed by at
least three anonymous referees which provided full written
evaluations.  Competition was high and after the first round of
refereeing only 25 full papers remained. Of these, 16 went through a
full second round of refereeing with written referee
reports. % , requested by the reviewers.
Finally, all 25 papers went through a final, copy-editing round. In
the end the special issue contains 17 technical papers, 6 application
papers, and 2 systems and tools papers.  During the first phase of
reviewing the papers submitted to the technical communications track
were also reviewed by at least three anonymous referees providing full
written evaluations. Also, a number of full paper submissions were
moved during the reviewing process to the technical communications
track.  Finally, 22 papers were accepted as technical
communications. A listing of these papers,
% collection of technical communications  
published in LIPIcs, appears at the end of the special issue.  The
list of the 25 accepted full papers, appearing in
this special issue, follows: \\

\newcommand{\tocTitle}[3]{\ \\ [-2.2mm] \href{#3}{#1}\\ [1mm]}
\newcommand{\tocAuthors}[1]{\hspace*{4mm}%
\begin{minipage}{0.9\textwidth}
\emph{#1}\\ [1.6mm]
\end{minipage}}
\newcommand{\tocFirstSection}[1]{\ \\ [-5mm] \textbf{#1}\\ [-4mm]}
\newcommand{\tocSecondSection}[1]{\ \\ [-4mm] \textbf{#1}\\ [-6mm]}

\tocFirstSection{Regular Papers}

% 20
\tocTitle{Automated Termination Analysis for Logic Programs with Cut}{363}
         {http://arxiv.org/abs/1007.4908}
    \tocAuthors{Peter Schneider-Kamp, J\"{u}rgen Giesl, Thomas
      Stroeder, Alexander  Serebrenik, Ren\'{e} Thiemann}
% 38
\tocTitle{Transformations of Logic Programs on Infinite Lists}{381}
         {http://arxiv.org/abs/1007.4157v1}
    \tocAuthors{Alberto Pettorossi, Maurizio Proietti, Valerio Senni}
% 90
\tocTitle{Swapping Evaluation: A Memory-Scalable Solution for
          Answer-On-Demand Tabling}{399} 
         {http://arxiv.org/abs/1007.3961}
    \tocAuthors{Pablo Chico de Guzm\'{a}n, Manuel Carro Li\~{n}ares, David S. Warren}
% 79
\tocTitle{Threads and Or-Parallelism Unified}{415}
         {http://arxiv.org/abs/1007.4438}
    \tocAuthors{V\'{i}tor Santos Costa, In\^{e}s Castro Dutra, Ricardo Rocha}
% 46
\tocTitle{CHR(PRISM)-based Probabilistic Logic Learning}{433}
         {http://arxiv.org/abs/1007.3858}
    \tocAuthors{Jon Sneyers, Wannes Meert, Joost Vennekens, Yoshitaka
      Kameya, Taisuke Sato} 
% 9
\tocTitle{Inference with Constrained Hidden Markov Models in PRISM}{449}
         {http://arxiv.org/abs/1007.5421}
    \tocAuthors{Henning Christiansen, Christian Theil Have, Ole Torp
      Lassen, Matthieu Petit} 
% 11
\tocTitle{A Translational Approach to Constraint Answer Set Solving}{465}
         {http://arxiv.org/abs/1007.4114}
    \tocAuthors{Christian Drescher, Toby Walsh}
% 13
\tocTitle{A Decidable Subclass of Finitary Programs}{483}
         {http://arxiv.org/abs/1007.3663} %***** 
    \tocAuthors{Sabrina Baselice, Piero Bonatti}
% 23
\tocTitle{Disjunctive ASP with Functions: Decidable Queries and
  Effective Computation}{499} 
         {http://arxiv.org/abs/1007.4028}
  \tocAuthors{Mario Alviano, Wolfgang Faber, Nicola Leone}
% 56
\tocTitle{Catching the Ouroboros: On Debugging Non-ground Answer-Set Programs}{517}
         {http://arxiv.org/abs/1007.4986}
    \tocAuthors{Johannes Oetsch, J\"{o}rg Puehrer, Hans Tompits}
% 19
\tocTitle{Loop Formulas for Description Logic Programs}{535}
         {http://arxiv.org/abs/1007.4040}
    \tocAuthors{Yisong Wang, Jia-Huai You, Li-Yan Yuan, Yi-Dong Shen}
% 55 
\tocTitle{Towards Closed World Reasoning in Dynamic Open Worlds}{551}
         {http://arxiv.org/abs/1007.4342}
    \tocAuthors{Martin Slota, Jo\~{a}o Leite}
% 102
\tocTitle{A Program-Level Approach to Revising Logic Programs under
  Answer Set Semantics}{569} 
         {http://arxiv.org/abs/1007.5024}
    \tocAuthors{James Delgrande}
% 61
\tocTitle{FO(FD): Extending classical logic with rule-based fixpoint definitions}{585}
         {http://arxiv.org/abs/1007.3819}
    \tocAuthors{Ping Hou, Broes De Cat, Marc Denecker}
% 16
\tocTitle{A Complete and Terminating Execution Model for Constraint Handling Rules}{601}
         {http://arxiv.org/abs/1007.3829}
    \tocAuthors{Hariolf Betz, Frank Raiser, Thom Fr\"{u}hwirth}
% 40
\tocTitle{Decidability Properties for Fragments of CHR}{617}
         {http://arxiv.org/abs/1007.4476}
    \tocAuthors{Maurizio Gabbrielli, Jacopo Mauro, Maria Chiara Meo, Jon Sneyers}
% 21
\tocTitle{A Declarative Semantics for CLP with Qualification and Proximity}{633}
         {http://arxiv.org/abs/1007.3629}
    \tocAuthors{Mario Rodr\'{\i}guez-Artalejo, Carlos A. Romero-D\'{\i}az}

\tocSecondSection{Application Papers and Systems and Tools Papers}

% 30
\tocTitle{Logic-Based Decision Support for Strategic Environmental Assessment}{651}
         {http://arxiv.org/abs/1007.3159}
    \tocAuthors{Marco Gavanelli, Fabrizio Riguzzi, Michela Milano,
      Paolo Cagnoli} % **************
% 83
\tocTitle{Test Case Generation for Object-Oriented Imperative Languages in CLP}{669}
         {http://arxiv.org/abs/1007.5195}
    \tocAuthors{Miguel G\'{o}mez-Zamalloa, Elvira Albert, Germ\'{a}n Puebla}
% 87
\tocTitle{Logic Programming for Finding Models in the Logics of
         Knowledge and its Applications: A Case Study}{687} 
         {http://arxiv.org/abs/1007.3700} % **********
    \tocAuthors{Chitta Baral, Gregory Gelfond, Enrico Pontelli, Tran
      Son} % ***************
% 72
\tocTitle{Applying Prolog to Develop Distributed Systems}{703}
         {http://arxiv.org/abs/1007.3835}
    \tocAuthors{Nuno P. Lopes, Juan Navarro Perez, Andrey Rybalchenko, Atul Singh}
% 29
\tocTitle{CLP-based Protein Fragment Assembly}{721}
         {http://arxiv.org/abs/1007.5180}
    \tocAuthors{Alessandro Dal Pal\`{u}, Agostino Dovier, Federico
      Fogolari, Enrico Pontelli} 
% 42 
\tocTitle{Formalization of Psychological Knowledge in Answer Set
  Programming and its Application}{737} 
         {http://arxiv.org/abs/1007.4767}
    \tocAuthors{Marcello Balduccini, Sara Girotto}
% 5
\tocTitle{Testing and Debugging Techniques for Answer Set Solver Development}{753}
         {http://arxiv.org/abs/1007.3223}
    \tocAuthors{Robert Brummayer, Matti J\"{a}rvisalo}
% 57
\tocTitle{The System Kato: Detecting Cases of Plagiarism for Answer-Set Programs}{771}
         {http://arxiv.org/abs/1007.4971}
    \tocAuthors{Johannes Oetsch, J\"{o}rg Puehrer, Martin Schwengerer, Hans Tompits}

%%% Local Variables: 
%%% mode: latex
%%% TeX-master: 00.tex
%%% End: 

\ \\ [-3mm]

We would like to thank very specially the members of the Program
Committee
% who took on their new role as co-guest editors, evaluating
% themselves or coordinating a good number of 
and the external referees
% , whom we very strongly thank as well, 
for their enthusiasm, hard work, and promptness, despite the higher
load of the two rounds of refereeing plus the copy editing phase. The
PC members were:
% required for journal publication:
%
Mar\'{\i}a Alpuente, % \emph{(Technical U. of Valencia, Spain)}, 
Pedro Cabalar, % \emph{(Coru\~{n}a University, Spain)}, 
Manuel Carro, % \emph{(Technical U.\ of Madrid, Spain)}, 
Luc De Raedt, % \emph{(K.\ U.\ Leuven, Belgium)}, 
Marina De Vos, % \emph{(University of Bath, UK)}, 
James Delgrande, % \emph{(Simon Fraser University, Canada)}, 
Marc Denecker, % \emph{(KU Leuven, Belgium)}, 
Agostino Dovier, % \emph{(University of Udine, Italy)}, 
Esra Erdem, % \emph{(Sabanci University, Istanbul, Turkey)}, 
Wolfgang Faber, % \emph{(University of Calabria, Italy)}, 
Thom Fruehwirth, % \emph{(University of Ulm, Germany)}, 
Maurizio Gabbrielli, % \emph{(University of Bologna, Italy)}, 
John Gallagher, % \emph{(Roskilde University, Denmark)}, 
Samir Genaim, % \emph{(Complutense University, Spain)}, 
Haifeng Guo, % \emph{(University of Nebraska at Omaha, USA)}, 
Joxan Jaffar, % \emph{(National U.\ of Singapore, Singapore)}, 
Tomi Janhunen, % \emph{(Helsinki U.\ of Technology, Finland)}, 
Michael Leuschel, % \emph{(U.\ of Duesseldorf, Germany)}, 
Alan Mycroft, % \emph{(U. of Cambridge, UK)}, 
Gopalan Nadathur, % \emph{(University of Minnesota, USA)}, 
Lee Naish, % \emph{(Melbourne University, Australia)}, 
Enrico Pontelli, % \emph{(New Mexico State University, USA)}, 
Vitor Santos Costa, % \emph{(University of Porto, Portugal)}, 
Tom Schrijvers, % \emph{(K.U. Leuven, Belgium)}, 
Tran Cao Son, % \emph{(New Mexico State University, USA)}, 
Peter J. Stuckey, % \emph{(Melbourne University, Australia)}, 
Terrance Swift, % \emph{(CENTRIA, Portugal)}, 
Peter Szeredi, % \emph{(Budapest U.\ of Tech.\ and E., Hungary)}, 
Frank Valencia, % \emph{(\'{E}cole Polytechnique, France)}, 
Wim Vanhoof, % \emph{(University of Namur, Belgium)}, 
Kewen Wang, % \emph{(Griffith University, Australia)}, 
Stefan Woltran, % \emph{(Vienna U.\ of Technology, Austria)}, 
and Neng-Fa Zhou. % \emph{(City University of New York, USA)}.

We would also like to thank David Basin, Francois
Fages, Deepak Kapur, and Molham Aref 
% for  agreeing to give giving invited talks, 
for their invited talks and those that helped organize ICLP: Veronica
Dahl % \emph{(Simon Fraser University, Canada)}
(General Chair and Workshops Chair), Marcello Balduccini
% \emph{(Kodak  Research Labs, USA)} 
and Alessandro Dal Pal\`{u} % \emph{(U. degli Studi di Parma, Italy)} 
(Doctoral Consortium), and Tom Schrijvers
% \emph{(K.U.\ Leuven, Belgium)} 
(Prolog Programming Contest).  
ICLP'10 was held as part of the 2010
Federated Logic Conference, hosted by the School of Informatics at the
% University 
U.\ of Edinburgh, Scotland.  Support by the conference sponsors
--EPSRC, NSF, Microsoft Research, Association for Symbolic Logic,
Google, HP, Intel-- is also gratefully acknowledged.  We are also
grateful to Andrei Voronkov for creating 
% (and helping us with) 
the EasyChair system.

Finally, we would like to thank very specially Ilkka Niemel\"{a},
editor in chief of Theory and Practice of Logic Programming, David
Tranah, from Cambridge University Press, Marc Herbstritt, from LIPIcs,
Leibniz Center for Informatics, 
all the members of the ALP Executive
Committee, and the ALP community in general for having believed in and
allowed us to put into practice this approach which we believe
provides compatibility with the publishing mechanisms of other fields
outside CS, without giving up the format and excitement
of our conferences. \\

\ \hfill Manuel Hermenegildo and Torsten Schaub

\ \hfill Program Committee Chairs and Guest Editors\\ [-5mm]

% --- Application and Systems Papers separate in table of contents!!!!!

\clearpage
\setcounter{page}{777}
\newcommand{\mytit}{Listing of the Technical Communications of the
  26th ICLP} \title{\mytit}
\ \\
\ \\ 
\noindent
{\Large \textbf{Listing of the Technical Communications of the
  26th ICLP}}\\

% % \newcommand{\tocTitle}[2]{\ \\ [-1.55mm] #1 \dotfill #2\\}
% \newcommand{\tocTitle}[2]{\ \\ [-1.5mm] #1 % \dotfill #2
%   \\}
% % \newcommand{\tocAuthors}[1]{{\raggedright \leftskip 15pt \rightskip 2.55em\itshape #1\endgraf}}
% \newcommand{\tocAuthors}[1]{\hspace*{4mm}\begin{minipage}{0.9\textwidth}
%   \emph{#1}
%   \end{minipage}\\}
% % \newcommand{\tocSection}[2]{\contentsline{chapter}{#1. \textmd{#2}}{0}}
% \newcommand{\tocSection}[1]{\ \\ [-2mm] {\large \textbf{#1}}\\ [-5mm]}

\ \\
This is a listing of the \emph{Invited Papers} and \emph{Technical
  Communications} which were also presented at ICLP 2010. These papers
constitute \href{http://www.dagstuhl.de/dagpub/978-3-939897-17-0}{Volume 7} of the
\href{http://www.dagstuhl.de/en/publications/lipics}{Leibniz
  International Proceedings in Informatics (LIPIcs)} series, published
on line through the
\href{http://drops.dagstuhl.de/}
{Dagstuhl Research Online Publication Server (DROPS).}  In addition to
these invited papers and technical communications the volume also
includes the 15 papers presented at the ICLP Doctoral Symposium.\\

\tocFirstSection{Preface}

  \tocTitle{Introduction to the Technical Communications of ICLP
    2010}{i}%{1}
           {http://dx.doi.org/10.4230/LIPIcs.ICLP.2010.XI}
    \tocAuthors{Manuel Hermenegildo and Torsten Schaub}

% \tocTitle{List of full papers (published in TPLP special
% issue)}{??}%{1}
%           {}

\tocSecondSection{Invited Papers}

  \tocTitle{Datalog for Enterprise Software: From Industrial
    Applications to Research (Invited Talk)}{1}%{1}
           {http://dx.doi.org/10.4230/LIPIcs.ICLP.2010.1}
    \tocAuthors{Molham Aref}
  \tocTitle{A Logical Paradigm for Systems Biology (Invited
    Talk)}{2}%{1}
           {http://dx.doi.org/10.4230/LIPIcs.ICLP.2010.2}
    \tocAuthors{Fran\c{c}ois Fages}

\tocSecondSection{Technical Communications}

  \tocTitle{Runtime Addition of Integrity Constraints in Abductive
    Logic Programs}{4}%{1}
           {http://dx.doi.org/10.4230/LIPIcs.ICLP.2010.4}
    \tocAuthors{Marco Alberti, Marco Gavanelli, Evelina Lamma}
  \tocTitle{Learning Domain-Specific Heuristics for Answer Set
    Solvers}{14}%{2}
           {http://dx.doi.org/10.4230/LIPIcs.ICLP.2010.14}
    \tocAuthors{Marcello Balduccini}
  \tocTitle{HEX Programs with Action Atoms}{24}%{3}
           {http://dx.doi.org/10.4230/LIPIcs.ICLP.2010.24}
    \tocAuthors{Selen Basol, Ozan Erdem, Michael Fink, Giovambattista Ianni}
  \tocTitle{Communicating Answer Set Programs}{34}%{4}
           {http://dx.doi.org/10.4230/LIPIcs.ICLP.2010.34}
    \tocAuthors{Kim Bauters, Jeroen Janssen, Steven Schockaert, Dirk Vermeir, Martine De Cock}
  \tocTitle{Implementation Alternatives for Bottom-Up
    Evaluation}{44}%{5}
           {http://dx.doi.org/10.4230/LIPIcs.ICLP.2010.44}
    \tocAuthors{Stefan Brass}
  \tocTitle{Inductive Logic Programming as Abductive Search}{54}%{6}
           {http://dx.doi.org/10.4230/LIPIcs.ICLP.2010.54}
    \tocAuthors{Domenico Corapi, Alessandra Russo, Emil Lupu}
  \tocTitle{Efficient Solving of Time-Dependent Answer Set
    Programs}{64}%{7}
           {http://dx.doi.org/10.4230/LIPIcs.ICLP.2010.64}
    \tocAuthors{Timur Fayruzov, Jeroen Janssen, Martine De Cock, Chris Cornelis, Dirk Vermeir}
  \tocTitle{Improving the Efficiency of Gibbs Sampling for
    Probabilistic Logical Models by Means of Program
    Specialization}{74}%{8} 
           {http://dx.doi.org/10.4230/LIPIcs.ICLP.2010.74}
    \tocAuthors{Daan Fierens}
  \tocTitle{Focused Proof Search for Linear Logic in the Calculus of
    Structures}{84}%{9}
           {http://dx.doi.org/10.4230/LIPIcs.ICLP.2010.84}
    \tocAuthors{Nicolas Guenot}
  \tocTitle{Sampler Programs: The Stable Model Semantics Abstract
    Constraint Programs Revisited}{94}%{10}
           {http://dx.doi.org/10.4230/LIPIcs.ICLP.2010.94}
    \tocAuthors{Tomi Janhunen}
  \tocTitle{A Framework for Verification and Debugging of Resource
    Usage Properties}{104}%{11}
           {http://dx.doi.org/10.4230/LIPIcs.ICLP.2010.104}
    \tocAuthors{Pedro Lopez-Garcia, Luthfi Darmawan, Francisco Bueno}
  \tocTitle{Contractible Approximations of Soft Global
    Constraints}{114}%{12}
           {http://dx.doi.org/10.4230/LIPIcs.ICLP.2010.114}
    \tocAuthors{Michael Maher}
  \tocTitle{Dedicated Tabling for a Probabilistic Setting}{124}%{13}
           {http://dx.doi.org/10.4230/LIPIcs.ICLP.2010.124}
    \tocAuthors{Theofrastos Mantadelis, Gerda Janssens}
  \tocTitle{Tight Semantics for Logic Programs}{134}%{14}
           {http://dx.doi.org/10.4230/LIPIcs.ICLP.2010.134}
    \tocAuthors{Luis Moniz Pereira, Alexandre Miguel Pinto}
  \tocTitle{From Relational Specifications to Logic Programs}{144}%{15}
           {http://dx.doi.org/10.4230/LIPIcs.ICLP.2010.144}
    \tocAuthors{Joseph Near}
  \tocTitle{Methods and Methodologies for Developing Answer-Set
    Programs---Project Description}{154}%{16}
           {http://dx.doi.org/10.4230/LIPIcs.ICLP.2010.154}
    \tocAuthors{Johannes Oetsch, Joerg Puehrer, Hans Tompits}
  \tocTitle{Tabling and Answer Subsumption for Reasoning on Logic
    Programs with Annotated Disjunctions}{162}%{17}
           {http://dx.doi.org/10.4230/LIPIcs.ICLP.2010.162}
    \tocAuthors{Fabrizio Riguzzi, Terrance Swift}
  \tocTitle{Subsumer: A Prolog Theta-Subsumption Engine}{172}%{18}
           {http://dx.doi.org/10.4230/LIPIcs.ICLP.2010.172}
    \tocAuthors{Jose Santos, Stephen Muggleton}
  \tocTitle{Using Generalized Annotated Programs to Solve Social
    Network Optimization Problems}{182}%{19} 
           {http://dx.doi.org/10.4230/LIPIcs.ICLP.2010.182}
    \tocAuthors{Paulo Shakarian, V.S. Subrahmanian, Maria Luisa Sapino}
  \tocTitle{Abductive Inference in Probabilistic Logic
    Programs}{192}%{20}
           {http://dx.doi.org/10.4230/LIPIcs.ICLP.2010.192}
    \tocAuthors{Gerardo Simari, V.S. Subrahmanian}
  \tocTitle{Circumscription and Projection as Primitives of Logic
    Programming}{202}%{21}
           {http://dx.doi.org/10.4230/LIPIcs.ICLP.2010.202}
    \tocAuthors{Christoph Wernhard}
  \tocTitle{Timed Definite Clause Omega-Grammars}{212}%{22}
           {http://dx.doi.org/10.4230/LIPIcs.ICLP.2010.212}
    \tocAuthors{Neda Saeedloei, Gopal Gupta}

\end{document}